\documentclass[letterpaper, 10 pt, conference]{ieeeconf}  

\IEEEoverridecommandlockouts                              

\usepackage{graphicx} 
\usepackage{amsmath} 
\usepackage{amssymb}  
\usepackage{subcaption}
\usepackage{siunitx}
\usepackage{cite}
\usepackage{bm}
\usepackage{mathtools}
\usepackage{pifont}
\usepackage[dvipsnames]{xcolor}
\usepackage[font=footnotesize]{caption}
\usepackage[abbreviations]{glossaries-extra}
\usepackage{booktabs}
\usepackage[colorlinks = true,
            linkcolor = blue,
            urlcolor  = blue,
            citecolor = blue,
            anchorcolor = blue]{hyperref}
\usepackage[capitalise]{cleveref}
\usepackage[table]{xcolor}
\usepackage{makecell}

\newcommand{\greencheck}{{\color{ForestGreen}\checkmark}}
\newcommand{\blueminus}{{\color{blue} $\sim$}}
\newcommand{\redcross}{{\color{purple} \ding{55}}}
\newcommand{\approptoinn}[2]{\mathrel{\vcenter{
  \offinterlineskip\halign{\hfil$##$\cr
    #1\propto\cr\noalign{\kern2pt}#1\sim\cr\noalign{\kern-2pt}}}}}

\newcommand{\appropto}{\mathpalette\approptoinn\relax}
\usepackage{float}
\usepackage{flushend}
\definecolor{earthgreen}{RGB}{230,255,230}
\definecolor{marsred}{RGB}{255,230,230}
\definecolor{moongrey}{RGB}{240,240,240}
\definecolor{superblue}{RGB}{210,240,255}

\title{Energy-Efficient Learning-Based Control of a Legged Robot \\ in  Multiple Gravity Environments}

\author{Philip Arm$^{\ast}$, Oliver Fischer$^{\ast}$, Joseph Church, Adrian Fuhrer, Hendrik Kolvenbach, and 
Marco Hutter
\thanks{$^{\ast}$These authors contributed equally to this work. All authors are with ETH Zurich, Robotics Systems Lab; Leonhardstrasse 21, 8092 Zurich, Switzerland.
        Contact: {\tt\small parm@ethz.ch}}%
\thanks{This research was supported by the Swiss National Science Foundation (SNF) through the National Centre of Competence in Digital Fabrication (NCCR dfab). LunarLeaper acknowledges funding through the CHiwi Foundation, SERI's NASO program (1-012571, LunarLeaper 2024), Moog Inc. and maxon.
}
}
\begin{document}

\bstctlcite{IEEEexample:BSTcontrol}

\maketitle
\thispagestyle{empty}
\pagestyle{empty}

\begin{abstract}

Legged robots are promising candidates for exploring challenging areas on low-gravity bodies such as the Moon, Mars, or asteroids, thanks to their advanced mobility on unstructured terrain. However, as planetary robots' power and thermal budgets are highly restricted, these robots need energy-efficient control approaches that easily transfer to multiple gravity environments. In this work, we introduce a reinforcement learning-based control approach for legged robots with gravity-scaled power-optimized reward functions. We use our approach to develop and validate a locomotion controller and a base pose controller in gravity environments from lunar gravity (\SI{1.62}{\meter \per \second^2}) to a hypothetical super-Earth (\SI{19.62}{\meter \per \second^2}) in simulation. Our approach successfully scales across these gravity levels for locomotion and base pose control with the gravity-scaled reward functions. On the real system, the power-optimized locomotion controller reached a power consumption for locomotion of \SI{23.4}{\watt} in Earth gravity on a \SI{15.65}{\kilogram} robot at \SI{0.4}{\meter \per \second}, a \SI{23}{\percent} improvement over a baseline policy. Additionally, we designed a constant-force spring offload system that allowed us to conduct real-world experiments on legged locomotion in lunar gravity. In this test setup, the power-optimized control policy reached \SI{12.2}{\watt}, \SI{36}{\percent} less than a baseline controller which is not optimized for power efficiency. Our method provides a scalable approach to developing power-efficient locomotion controllers for legged robots across multiple gravity levels.
\end{abstract}


\section{INTRODUCTION}
The robotic exploration of our solar system is crucial to advancing our understanding of the solar system and paving the way for future human missions to other planetary bodies. Although orbiters and stationary landers can provide an initial insight into the conditions on other planetary bodies, robotic surface missions are essential to answer more scientific questions and prepare for human exploration. Many targets of interest, such as permanently shadowed regions close to the lunar south pole, crater walls, and lava tubes, are difficult to access. Therefore, we require robots with advanced mobility to venture into these challenging environments.

To date, rovers used for planetary exploration have almost exclusively relied on wheeled locomotion, starting from the first rover - Lunokhod 1~\cite{kassel1971lunokhod} - to Mars rovers~\cite{lindemann2006mars, grotzinger2012mars, farley2020mars}, and modern Moon rovers~\cite{ding20222, mathavaraj2020isro}. Accordingly, wheeled systems can build on decades of flight heritage and provide reliable mobility in relatively flat terrain. However, they reach their limits on steep slopes with granular terrain and unstructured, highly uncertain environments~\cite{david2005opportunity}. 

In the meantime, legged robots have shown impressive locomotion capabilities on Earth in the last decade. Powered by recent advances in reinforcement learning-based control, these robots can now explore unstructured natural terrain~\cite{lee2020learning, miki2022learning, agha2021nebula} and overcome obstacles higher than their body height~\cite{hoeller2023anymal, cheng2024extreme}.
\begin{figure}[t]
    \centering
    \includegraphics[width=1\columnwidth]{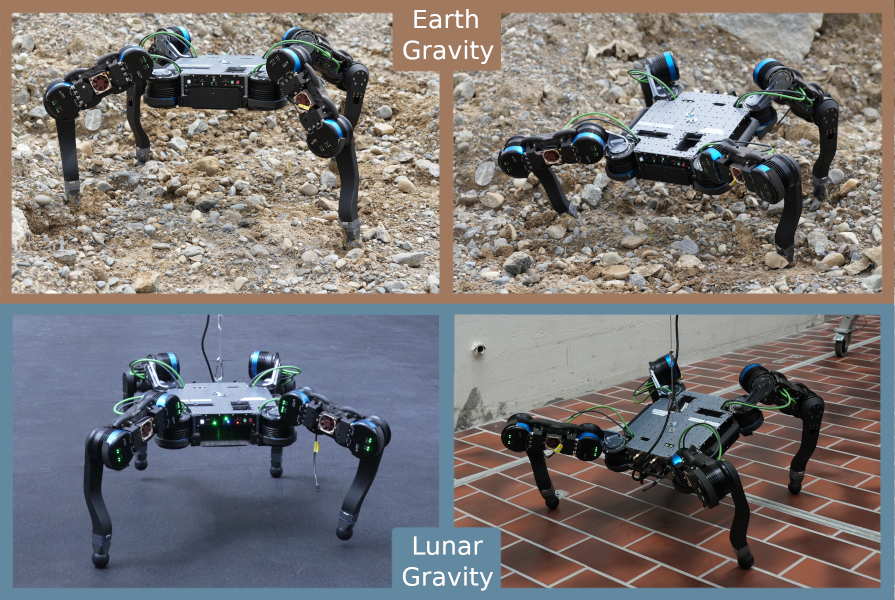}
    \caption{Our robot can efficiently walk in earth gravity (top left) and in a lunar gravity test setup (bottom left) using gravity-scaled power-optimized control policies. Additionally, we created a base pose tracking controller that can operate in the same environments (top and bottom right). }
    \label{fig:eyecatcher}
\end{figure}
Multiple groups have investigated the use of legged robots for planetary exploration~\cite{dirk2007bio,bartsch2012spaceclimber,roennau2014reactive, arm2019spacebok}. Today, legged robots have already been used successfully in analog mission scenarios such as the ESA-ESRIC Space Resources Challenge 2021-2022~\cite{schnell2023efficient, arm2023scientific} and the NASA BRAILLE project~\cite{morrell2024robotic}.

Based on these recent successes and considering the rugged terrain surrounding lunar pits, we selected a legged robot for the LunarLeaper mission \footnote{https://www.lunarleaper.space/} to explore lava tubes on the lunar surface~\cite{kolvenbach2024lunarleaper}. In addition to advanced locomotion capabilities, a legged robot offers additional benefits when operating base-mounted instruments, as it can control its base pose while standing. This feature is useful, for example, for camera tilting, antenna alignment, and orienting solar panels relative to the Sun. However, while legged systems have proven to operate well in steep terrain and provide value in analog missions, we still lack validation of reinforcement learning-based control of legged robots in lunar gravity. Furthermore, due to the power and thermal constraints on the lunar surface, we need to focus more on power-efficient control of legged robots.

Consequently, this paper focuses on power-efficient reinforcement learning-based control of legged robots across multiple tasks and gravity environments. While we do develop our controllers to work on rough terrain, we limit our evaluations to flat, rigid ground for comparability. Specifically, our contributions are the following:
\begin{itemize}
    \item We show that scaling reward functions with gravity in a physically meaningful manner produces working controllers in different gravity environments for multiple tasks.
    \item We compare baseline and power-optimized reward functions for our controllers in different gravity levels and validate that using power-optimized reward functions provides high-efficiency locomotion controllers in multiple gravity levels, both in simulation and on the real robot.
\end{itemize}

\begin{figure*}[h!]
    \centering
    \includegraphics[width=\textwidth]{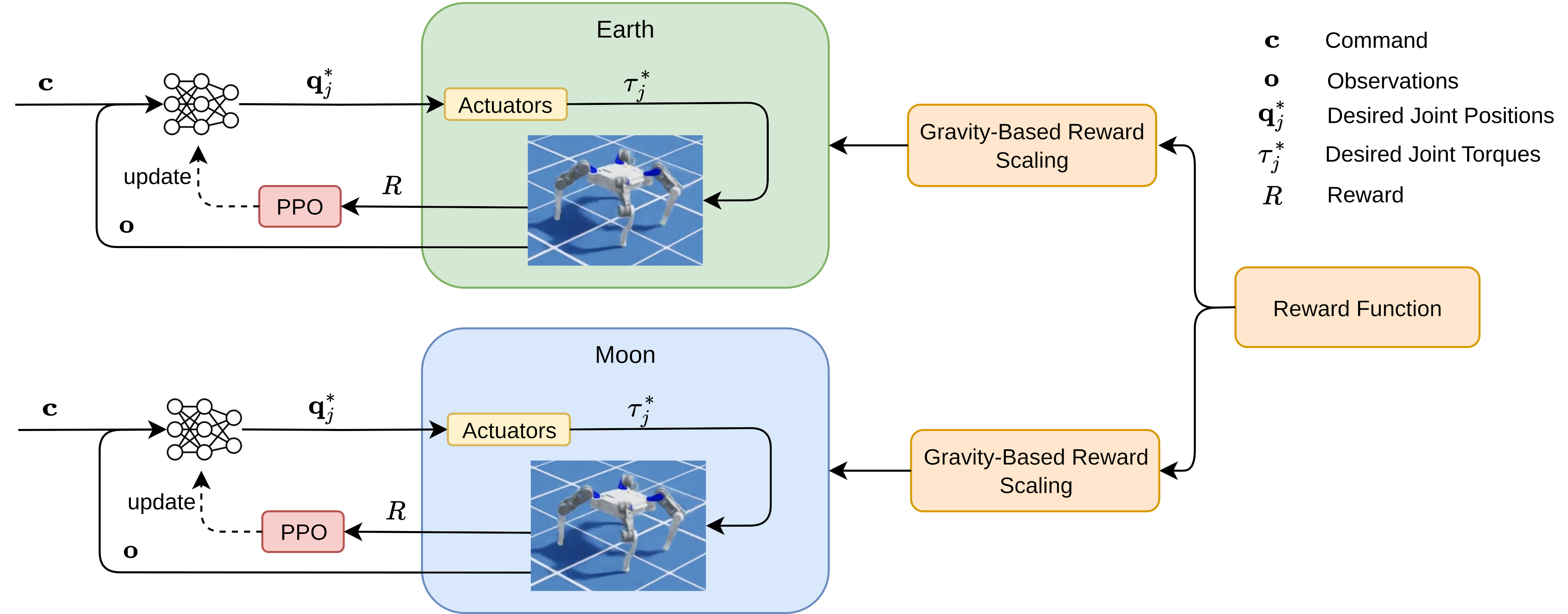}
    \caption{\textbf{Overview of our training setup}: We use PPO~\cite{schulman2017proximal} with an asymmetric actor-critic setup. Scaling the reward function with gravity based on first principles allows our setup to scale across multiple gravity levels.}
    \label{fig:method_overview}
\end{figure*}

\section{RELATED WORK}

\subsection{Legged Locomotion in Space}
Multiple research projects have evaluated legged robots in analog locomotion tests on steep slopes and granular soil~\cite{dirk2007bio, bartsch2012spaceclimber, kolvenbach2021martianslopes}. These works have shown that legged robots can traverse slopes up to \SI{35}{\degree} on firm ground~\cite{bartsch2012spaceclimber} and \SI{25}{\degree} on Martian analog soil~\cite{kolvenbach2021martianslopes}. Some groups even built free-climbing robots capable of climbing vertical walls~\cite{parness_lemur_2017, tanaka2022scaler}. Although these works showed impressive performance on challenging terrain, none of them considered the reduced gravity in planetary exploration.

Other works focused on legged mobility in low- and microgravity environments~\cite{rudin2021catlike,olsen2023design, spiridonov2024spacehopper,kolvenbach2018isairas}. However, these works are either limited to simulation experiments or validated the system in reduced dimensionality in microgravity. None of these works investigated and evaluated legged locomotion on a real robot in lunar gravity.

\subsection{Power-Efficient Legged Locomotion}
Multiple works have investigated the development of more power-efficient locomotion policies using deep reinforcement learning. In most of the works, power consumption is calculated in simulation based on joint torque and velocity values and penalized during training~\cite{fu2021minimizingenergyconsumptionleads,yang2021fastefficientlocomotionlearned}. \textit{Valsecchi et al.}~\cite{valsecchi2024accurate}  trained a neural network to directly estimate energy consumption given the joint states based on data collected on the real robot. This data-driven power model was then used to estimate and penalize power consumption during training. \textit{Mahankali et al.} optimized power consumption as an auxiliary objective under the constraint that the policy must obtain the maximum task reward that a policy trained without regularization could achieve~\cite{mahankali_maximizing_2024}.
While these approaches for efficient locomotion have been extensively validated in Earth's gravity, a validation of these power-based approaches across different gravity levels is still missing.

\section{METHOD}
We implemented our controllers on an improved version of the \textit{Magnecko} quadrupedal robot~\cite{leuthard2025magnecko}. \textit{Magnecko} is a \SI{15.65}{\kilogram} robot in insect-style configuration with a leg length of \SI{0.5}{\meter} (Fig.~\ref{fig:eyecatcher}).

We build on modern reinforcement learning pipelines to train control policies for two tasks:  A locomotion control policy, tracking planar velocity commands, and a base pose control policy, tracking the base height, pitch, and yaw. These policies allow us to validate the gravity scaling of the rewards~(Sec.~\ref{sec:gravity_scaling}) and the power consumption of baseline and power-optimized policies in different gravity levels~(Sec.~\ref{sec:power_reward}). Section \ref{sec:mdp} details the training setup of the two policies.

We use IsaacLab as a simulation environment~\cite{IsaacLab} and Proximal Policy Optimization (PPO) as the deep reinforcement learning algorithm~\cite{schulman2017proximal} with multi-level perceptrons for actor and critic networks with hidden layer sizes \{512, 256, 128\}. We identified joint friction, damping, and armature using the system identification method proposed by \textit{Bjelonic et al.}~\cite{bjelonic2025towards} for effective sim-to-real transfer. We trained our policies for 3000 iterations on a RTX4090 GPU, which takes \SI{3}{\hour} per training.

\subsection{Gravity-Based Reward Scaling}
\label{sec:gravity_scaling}
To design a gravity-based reward scaling, we look at the equation of motion of a multi-body system
\begin{align*}
&\bm{\tau} = \bm{M}(\bm{q}) \ddot{\bm{q}} + \bm{b}(\bm{q}, \dot{\bm{q}}) + \bm{g}(\bm{q}) - \bm{J}_c(\bm{q})^T \bm{F}_c  \quad , \\
&\text{where} \\
\bm{\tau} \quad &\text{Torques} \\
\bm{M}(\bm{q}) \quad&\text{Mass matrix} \\
\bm{q}, \dot{\bm{q}}, \ddot{\bm{q}} \quad  &\text{Position, velocity and acceleration vectors} \\
\bm{b}(\bm{q}, \dot{\bm{q}}) \quad &\text{Coriolis and centrifugal terms} \\
\bm{g}(\bm{q}) \quad &\text{Gravitational terms} \\
\bm{J}_c(\bm{q}) \quad &\text{Jacobian corresponding to the contact forces} \\
\bm{F}_c \quad &\text{Contact forces} \quad . \\
\end{align*}
As \SI{67}{\percent} of \textit{Magnecko's} mass is located in the base and the hip actuators, we assume that the inertial forces produced by the legs are low. Furthermore, we assume that the base does not experience high accelerations, so the contact forces $\bm{F}_c$ are dominated by the gravity force $\bm{F}_g$. As a first-order approximation for the gravity scaling, we accordingly assume that the torques are approximately proportional to the gravity
\begin{equation*}
    \bm{\tau} \appropto g \quad .
\end{equation*}

We define the default reward weight in Earth's gravity as $w_{iE}$ for reward function $i$ and the gravity factor $\alpha_g = \frac{g_E}{g}$, where $g$ is the target gravity and $g_E$ is Earth's gravity. We then scale the weight $w_{iE}$ with the gravity factor $\alpha_g$ depending on the reward function's relation to the joint torques to preserve the relative magnitude of different reward functions across gravity environments. For example, for a reward function penalizing the sum of joint powers $\bm{\dot{q}} \bm{\tau}$, the function approximately scales with gravity as
\begin{equation*}
    \bm{\dot{q}} \bm{\tau} \propto \bm{\tau} \appropto g \quad.
\end{equation*}
Therefore, the weight of the respective reward function will be $w_i = \alpha_g w_{iE}$. Similarly, a reward function for the squared torque scales as
\begin{equation*}
    \bm{\tau}^2 \appropto g^2 \quad ,
\end{equation*}
and we, therefore, scale the respective weight as 
\begin{equation}\label{eq:rescaling_law}
    w_{i} = \alpha_g^2 w_{iE}
\end{equation}
\subsection{Power-Optimized Regularization Reward}
\label{sec:power_reward}
\begin{figure}[ht]
    \centering
    \includegraphics[width=1\columnwidth]{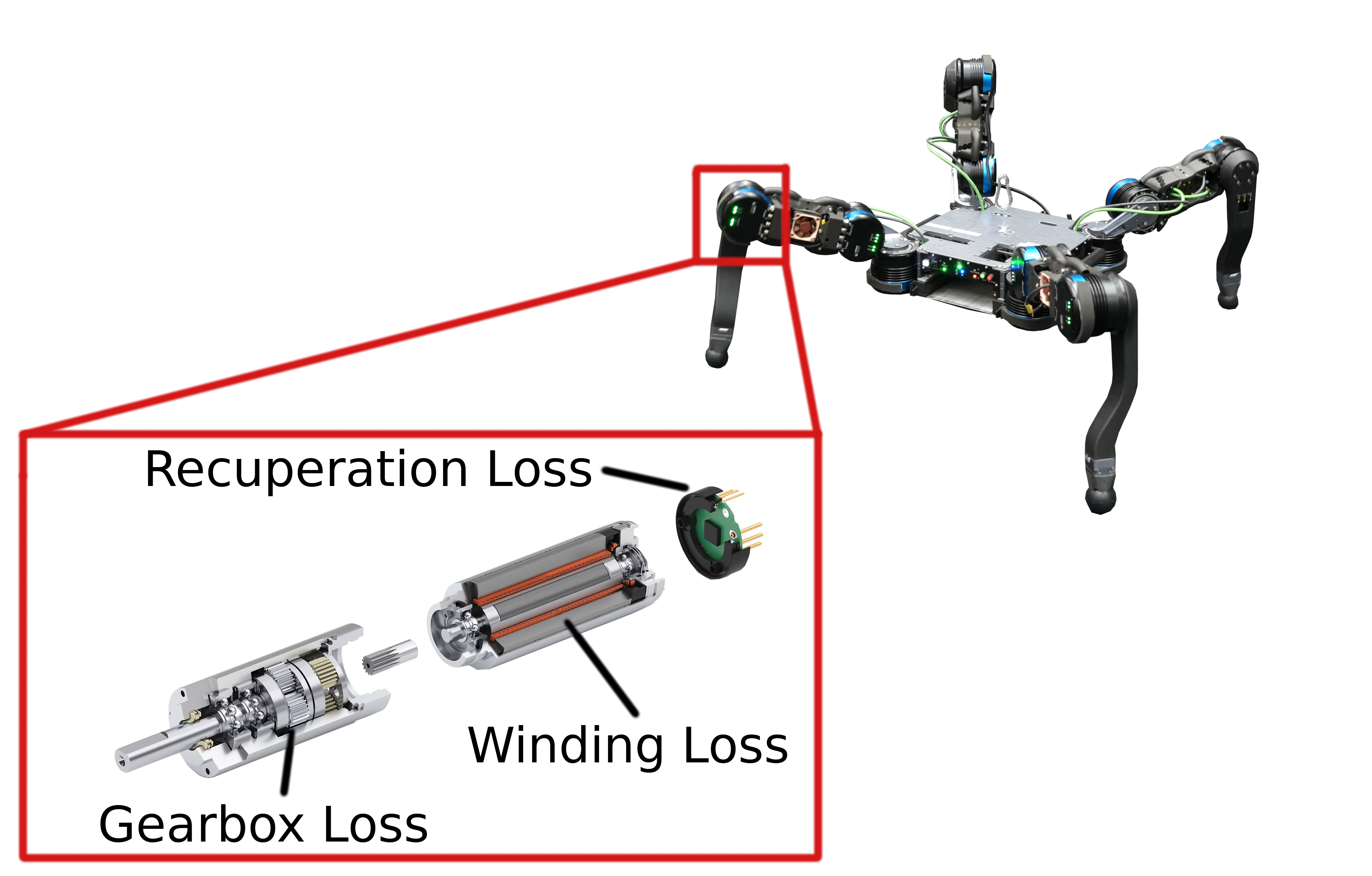}
    \caption{\textbf{Power losses in the drivetrain}. Recuperation loss occurs because not all the kinetic energy is effectively converted to electrical energy when the drivetrain is braking.}
    \label{fig:powertrain_losses}
\end{figure}

Aside from defining a scaling law for reward functions in different gravity levels, we require a reward function that minimizes the robot's power consumption while tracking the desired command. Previous work has shown that energy-based rewards generate robust and efficient gaits \cite{fu2021minimizingenergyconsumptionleads, mahankali_maximizing_2024}. Here, we model the total power loss as the sum of the recuperation loss and the winding loss (Fig.~\ref{fig:powertrain_losses}). The gearbox loss is implicitly included via the damping parameters we set in simulation based on the system identification pipeline~\cite{bjelonic2025towards}. Accordingly, we do not model it separately and formulate the power loss reward function as
\begin{equation}\label{eq:energy_penalty}
    r_{Power} = r_{Joint} + r_{Winding} .
\end{equation}
The recuperation loss is calculated as
\begin{equation}\label{eq:energ_penalty:power}
\begin{split}
    r_{\text{Joint}} =\ 
    &\max\left(\bm{\tau} \cdot \bm{\dot{q}},\ 0\right) \\
    &- \min\left(\eta_{\text{Recup}} \cdot \bm{\tau} \cdot \bm{\dot{q}},\ 0\right) \quad .
\end{split}
\end{equation}
\textit{Magnecko's} power electronics are not designed to recuperate power to the battery, so we set the recuperation parameter $\eta_{Recup}=0$ in this work. The winding loss is calculated as 
\begin{equation}\label{eq:energy_penalty:winding}
    r_{Winding} = \left( \frac{\bm{\tau}}{G \cdot k_t} \right)^2 \cdot R
\end{equation}
where $G$ is the gearbox ratio, $k_t$ is the torque constant, and $R$ is the winding resistance. The winding loss is dissipated because of resistive heating in the windings. We choose ${-3\cdot10^{-3}}$ as a weight for the power reward and apply the reward scaling separately for the two terms $r_{Joint}$ and $r_{Winding}$. $r_{\text{Joint}}$ is scaled with $\alpha_g$ and $r_{Winding}$ is scaled with $\alpha_g^2$.
\subsection{MDP Formulation for Locomotion and Base Pose Control}
\label{sec:mdp}
We trained controllers for locomotion control and base pose control to assess the performance of our gravity-based scaling and power-optimized rewards. For the locomotion controller, the command is defined as the base's $x$, $y$ and yaw velocity in base frame $\bm{c}_{loco} = \left ( \prescript{}{\mathcal{B}}{v_{Bx}}, \prescript{}{\mathcal{B}}{v_{By}}, \prescript{}{\mathcal{B}}{\dot \psi_{B}} \right )^T$. The command for the base pose tracking controller is the desired base height, pitch angle, and the yaw velocity in the local terrain frame $\bm{c}_{base} = \left ( \prescript{}{\mathcal{T}}{r_{Bz}}, \prescript{}{\mathcal{T}}{\theta_{B}}, \prescript{}{\mathcal{T}}{\dot \psi_{B}} \right) ^T $.
The actions $\bm{a} \in \mathbb{R}^{12}$ control the target joint positions for the proportional-derivative (PD) controllers of the robot's joints, and are updated at \SI{50}{Hz} The target positions $\bm{q}_*$ are computed as
\begin{equation}\label{eq:action_to_jointpos}
    \bm{q}_* = \sigma_a \cdot \bm{a} + \bm{q}_{def}
\end{equation}
where $\sigma_a = 0.3$ is the heuristically chosen action scaling factor, and $\bm{q}_{def} \in \mathbb{R}^{12}$ are the default joint positions. 
Aside from the different command formulations, the two controllers share the same observations. The actor observations $\bm{o}_a$ consist of the command $\bm{c}_{task}\in \mathbb{R}^3$, the IMU angular velocity readings at the last eight time steps $\bm{\omega}_{[t,\dots,t-8]} \in \mathbb{R}^{24}$, the gravity vector from the IMU projected into the base frame $\bm{g}_p \in \mathbb{R}^3$, the joint positions $\bm{q} \in \mathbb{R}^{12}$, the joint velocities $\bm{\dot{q}} \in \mathbb{R}^{12}$, and the previous actions $\bm{a}_{t-1} \in \mathbb{R}^{12}$:
\begin{equation}\label{eq:actor_obs}
    \bm{o}_a = \left[\bm{c}_{task}, \bm{\omega}_{[t,\dots,t-8]}, \bm{g}_{p}, \bm{q}, \bm{\dot{q}}, \bm{a}_{t-1}\right] \in \mathbb{R}^{66}
\end{equation}
The critic observations $\bm{o}_c$ are largely the same as the actor observations, with the only difference being that instead of receiving the IMU history $\bm{\omega}_{[t,\dots,t-8]}$, the critic receives the ground truth base twist $\bm{\xi} \in \mathbb{R}^6$, making the observation
\begin{equation}\label{eq:critic_obs}
    \bm{o}_c = [\bm{c}_{task}, \bm{\xi}, \bm{g}_p, \bm{q}, \bm{\dot{q}}, \bm{a}_{t-1}] \in \mathbb{R}^{48}
\end{equation}

Tab.~\ref{tab:rewards} shows all rewards used in this work. Both controllers share a set of base regularizations. Each controller also has separate reward functions to track the respective command and provide additional task-specific regularizations. We trained each controller twice, with two separate sets of general regularizations. The "baseline" consists of commonly used regularizations in legged locomotion controllers~\cite{rudin2022learning}. The "power-optimized" set uses the power penalty described in Eq.~\ref{eq:energy_penalty}. While the power penalty has only a single weight in the formulation for Earth gravity, we rescale the recuperation losses and the winding losses separately according to Eq.~\ref{eq:rescaling_law}.

As the robot has two symmetry axes (left/right and front/back), we provide symmetry data augmentation according to \textit{Mittal et al.}~\cite{mittal2024symmetry}. We add disturbances to the robot's base in simulation by applying randomized velocities of up to \SI{0.7}{\meter \per \second} and constant forces of up to \SI{5}{\newton}. The robot's mass is randomized by \SI{\pm 2}{\kilogram}. The static friction coefficient is randomized between $0.4$ and $1.4$, and the dynamic friction coefficient is randomized between $0.1$ and $1.4$. We use the same terrain set for both controllers, consisting of a mix of rough terrains: Slopes of up to \SI{23}{\degree}, random boxes of up to \SI{0.1}{m}, and noise terrain of up to \SI{0.1}{m}.

We add a curriculum for the terrain~\cite{rudin2022learning} and the power penalty, which allows the controller to first learn the tracking behavior under easy conditions and only then increase the terrain difficulty and optimize for efficient behavior.

As we test the policies on a gravity offset rig in Earth gravity (Sec.~\ref{sec:gravity_offload}), the legs will experience Earth gravity during deployment. To still allow sim-to-real transfer of policies trained in low gravity, we pass the gravity compensation torques $\bm{g}(\bm{q})$ of the legs as feedforward terms to the actuators both in training and deployment. Accordingly, the legs are always gravity compensated, and the policies only need to regulate around this gravity compensation.

\begin{table}[h]
    \centering
    \begin{tabular}{lcr}
        \toprule
        \textbf{Term} & \textbf{Equation} & \textbf{Weight} \\
        \midrule \midrule
        \multicolumn{3}{l}{\textbf{Shared Regularizations}} \\
        \midrule
        Joint limits & \makecell[{{l}}]{\renewcommand{\arraystretch}{2.0}$\sum_{i=1}^{12} \max\left(q_i-q_{max,i},0\right)$ \\ $-\min\left(q_i - q_{min,i}, 0\right)$} & $-2.0$ \\
        Undesired contacts & $ n_c $ & $-2.0$ \\
        \midrule
        \multicolumn{3}{l}{\textbf{Baseline Regularizations}} \\
        \midrule
        Torque & $\left|\bm{\tau}\right|^2$ & $-\alpha_g^2 \cdot 10^{-4}$ \\
        Action rate & $ \left| \bm{a}_{t} - \bm{a}_{t-1} \right|^2 $ & $-0.08$ \\
        Joint acceleration & $ \left| \bm{\ddot{q}} \right|^2 $ & $-8 \cdot 10^{-7}$ \\
        \midrule
        \multicolumn{3}{l}{\textbf{Power Regularizations}} \\
        \midrule
        Power & $r_{\text{Power}}$ (see Eq.~\ref{eq:energy_penalty}) & \cref{sec:power_reward} \\
        \midrule
        \multicolumn{3}{l}{\textbf{Locomotion Rewards}} \\
        \midrule
        Linear tracking & $ \exp\left\{ -\left(e_{v,x}^2 + e_{v,y}^2\right)/0.25 \right\} $ & $1.0$ \\
        Yaw tracking & $ \exp\left\{- e_{\psi}^2 / 0.25 \right\} $ & $0.5$ \\
        Foot impact velocity & $ \left| \prescript{}{\mathcal{F}}{\bm{v}_{F}} \right|^2 $ (on contact) & $-0.6$ \\
        \midrule
        \multicolumn{3}{l}{\textbf{Base Pose Rewards}} \\
        \midrule
        Height tracking & $ \exp\left\{ - e_h^2 / 0.1^2 \right\} $ & $1.0$ \\
        Pitch tracking & $ \exp\left\{ - e_{\theta}^2 / 0.15^2 \right\} $ & $1.0$ \\
        Yaw Tracking & $ \exp\left\{ - e_{\dot{\psi}}^2 / 0.15^2 \right\} $ & $2.0$ \\
        XY Velocity & $ \left| v_x^2 + v_y^2 \right| $ & $-0.6$ \\
        \bottomrule
    \end{tabular}
    \caption{Rewards used to train the locomotion and base pose controllers. In total, four policies are trained for each gravity by combining different task and regularization rewards. Command tracking errors are denoted according to the scheme $e_{v,x}=c_{v,x} - \prescript{}{\mathcal{B}}{v_x}$. $\alpha_g = \frac{g_E}{g}$ denotes the gravity scaling factor. $i$ denotes the $i^{th}$ joint.}
    \label{tab:rewards}
\end{table}

\subsection{Gravity Offload System}

\begin{figure*}[h!]
  \begin{minipage}{0.48\textwidth}
    \includegraphics[width = \textwidth]{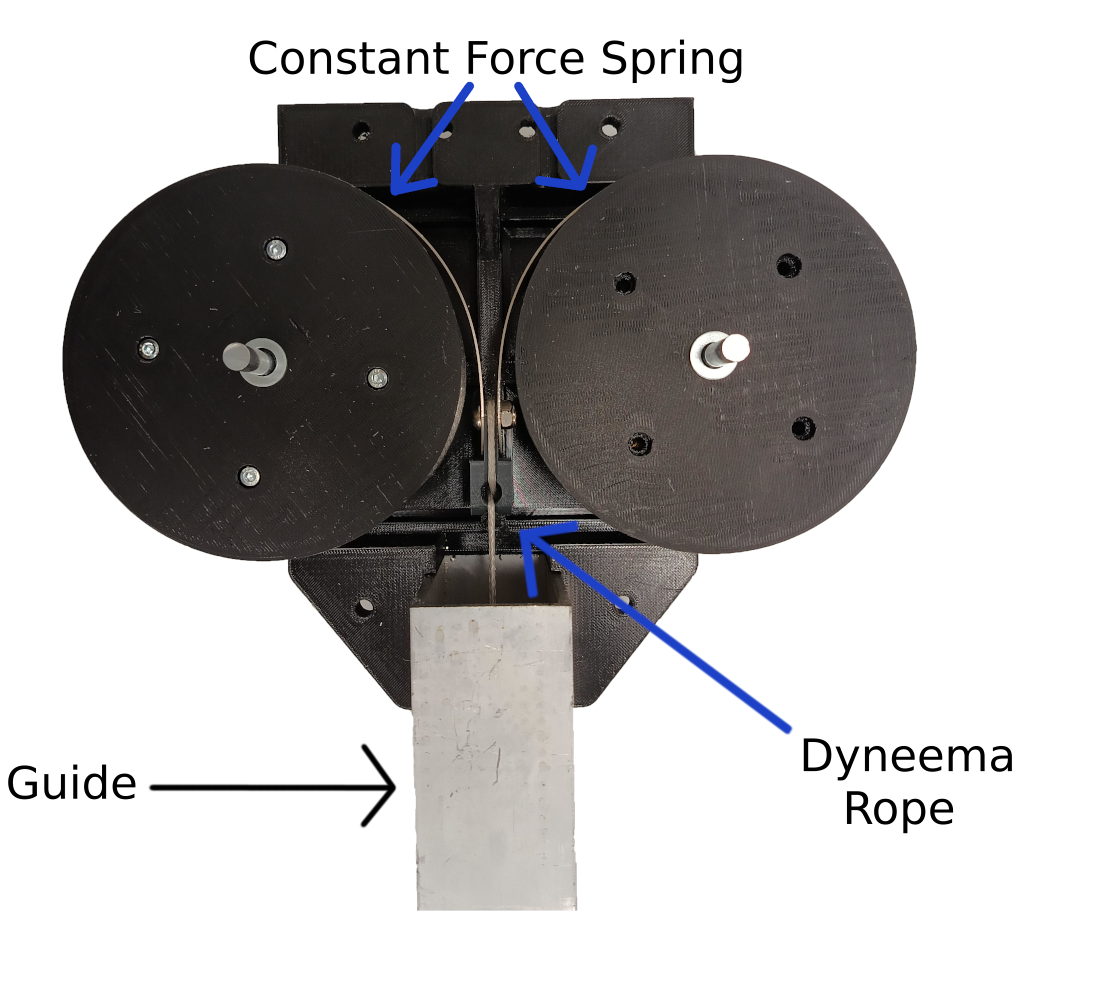}
  \end{minipage}
  \hfill
  \begin{minipage}{0.48\textwidth}
    \includegraphics[width = \textwidth]{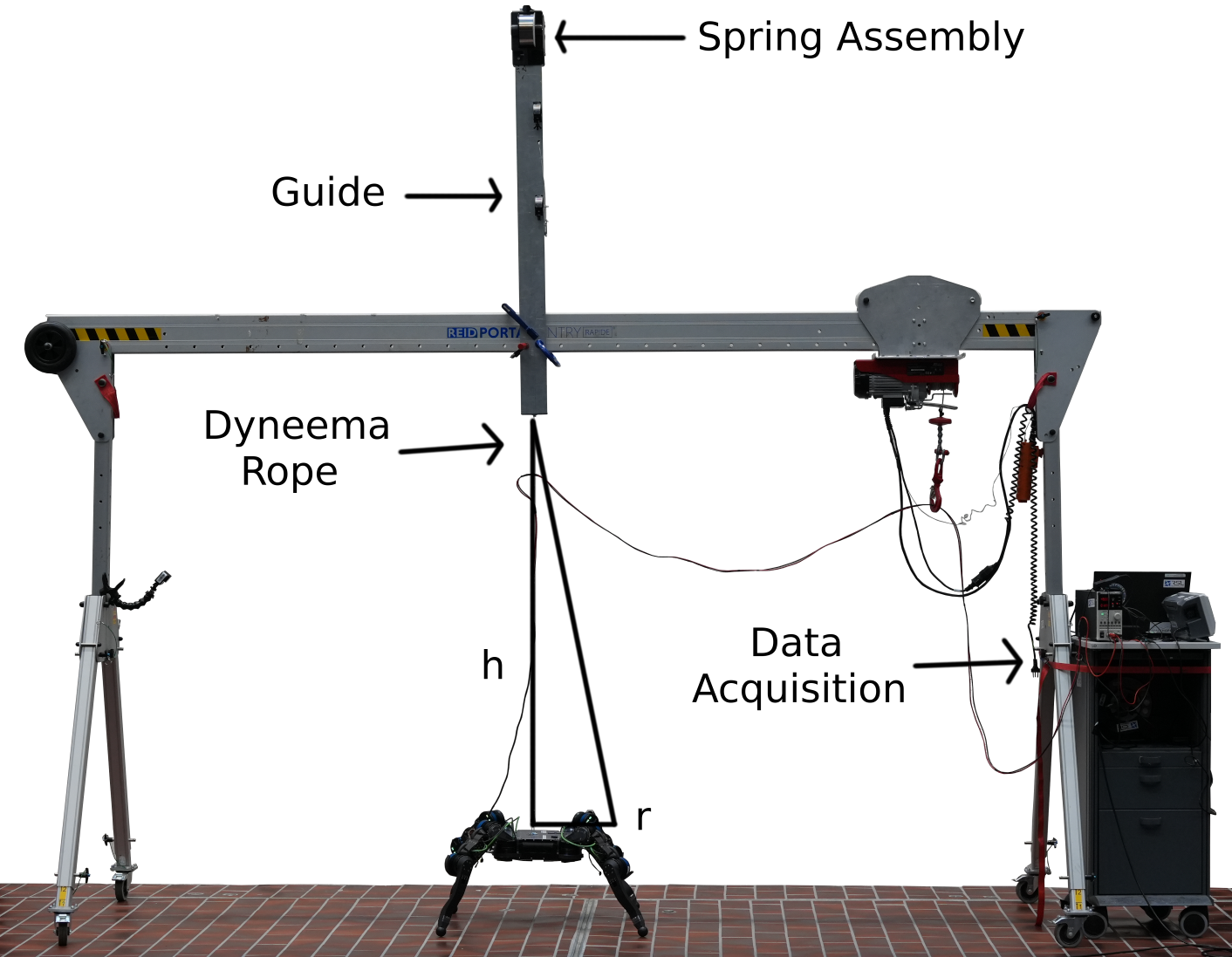}
  \end{minipage}
  \caption{Our constant force spring offload system (left) allows us to test lunar locomotion policies on the real robot. We mounted the system on a wheeled gantry (right) to conduct locomotion tests and measure the robot's power consumption during locomotion.}
  \label{fig:test_setup}
\end{figure*}
\label{sec:gravity_offload}

We developed a passive gravity-offload system to validate our reinforcement learning controllers in the real world in simulated lunar gravity. The system consists of two constant-force spring coils (Fig.~\ref{fig:test_setup}). The ends of the springs connect to a \SI{2}{\milli\meter} Dyneema rope. We chose a passive spring-loaded system to avoid the added inertia of a counterweight system and the control bandwidth limitations of an active system. Our system adds approximately \SI{1.5}{\kilogram} of parasitic mass compared to the \SI{13}{\kilogram} required for a simple counterweight. 

The rope is routed through an aluminum extrusion to ensure the springs are always fully enclosed, even at full extension. At the end of the extrusion, the rope exits through a PTFE guide to minimize friction. We attached the rope to the robot's base near the center of gravity.

To measure the actual offload force of the system, we measured the robot's mass on a scale and determined the offload force to be \SI{117.2}{\newton}, which is lower than the required vertical offload force of \SI{128.2}{\newton} for \textit{Magnecko}, most likely due to manufacturing inaccuracies in the spring system. To compensate for this error, we removed the \SI{0.8}{\kilogram} battery from the robot for the lunar gravity tests, which reduces the gravity force by \SI{7.9}{\newton}. We accordingly shifted the attachment point of the offload mechanism to be close to the new center of mass of the robot. Additionally, we increased the robot's mass by \SI{1.9}{\kilogram} in simulation for the lunar policy training, compensating the residual \SI{3.1}{\newton}.

We mounted the offload system on a wheeled gantry to follow the robot in the testing area and keep the Dyneema rope close to vertical at all times (Fig.~\ref{fig:test_setup}). In this setup, we introduce an additional error source when the rope is not vertical. The respective errors in the vertical and horizontal directions are
\begin{equation}
    F_{\epsilon,z}(r) = F_{Spring}\left(1-\frac{h}{\sqrt{r^2+h^2}}\right)
\end{equation}
\begin{equation}
    F_{\epsilon,r}(r) = F_{Spring}\frac{r}{\sqrt{r^2+h^2}} \quad ,
\end{equation}

where $r$ is the radial deflection from the vertical axis. Assuming that $r <$ \SI{0.15}{\meter}, and given that the system is mounted at a height of \SI{1.9}{\meter}, the resulting vertical error is below \SI{0.37}{\newton}. In the radial direction, the maximum error force is \SI{9.27}{\newton}.

\section{RESULTS}

\subsection{Adapting to Multiple Gravity Environments with Gravity-Based Reward Scaling}

\begin{table*}[ht]
    \centering
    \begin{tabular}{c|c|c|c|c}
         \toprule
         & \textbf{Baseline} & \textbf{Gravity-Scaled Baseline} & \textbf{Power-Optimized} & \textbf{Gravity-Scaled \& Power-Optimized} \\
         \midrule
         \midrule
         \multicolumn{5}{c}{\textbf{Locomotion Task -  Qualitative Evaluation}} \\
         \midrule
         \rowcolor{earthgreen}
         Earth $\left(\SI{9.81}{\meter \per \second^2}\right)$ & \greencheck & \greencheck & \greencheck & \greencheck \\
         \rowcolor{marsred}
         Mars $\left(\SI{3.73}{\meter \per \second^2}\right)$ & \greencheck & \greencheck & \blueminus & \greencheck \\
         \rowcolor{moongrey}
         Moon $\left(\SI{1.62}{\meter \per \second^2}\right)$ & \blueminus & \blueminus & \redcross & \greencheck \\
        \rowcolor{superblue}
        Super-Earth $\left(\SI{19.62}{\meter \per \second^2}\right)$ & \greencheck & \greencheck & \greencheck & \greencheck \\
        \midrule
         \multicolumn{5}{c}{\textbf{Locomotion Task - Power Consumption (W)}} \\
         \midrule
         \rowcolor{earthgreen}
         Earth $\left(\SI{9.81}{\meter \per \second^2}\right)$ & 66.6 & 66.6 & \textbf{39.9} & \textbf{39.9} \\
         \rowcolor{marsred}
         Mars $\left(\SI{3.73}{\meter \per \second^2}\right)$ & 18.9 & 18.1 & 59.6 & \textbf{15.4} \\
         \rowcolor{moongrey}
         Moon $\left(\SI{1.62}{\meter \per \second^2}\right)$ & 10.6 & 8.3 & 26.5 & \textbf{3.93} \\
        \rowcolor{superblue}
         Super-Earth $\left(\SI{19.62}{\meter \per \second^2}\right)$ & 121.7 & 161.7 & \textbf{45.7}  & 172.3 \\
        \midrule
        \midrule
         \multicolumn{5}{c}{\textbf{Base Pose Task -  Qualitative Evaluation}} \\
         \midrule
         \rowcolor{earthgreen}
         Earth $\left(\SI{9.81}{\meter \per \second^2}\right)$ & \blueminus & \blueminus & \greencheck & \greencheck \\
         \rowcolor{marsred}
         Mars $\left(\SI{3.73}{\meter \per \second^2}\right)$ & \redcross & \blueminus & \blueminus & \blueminus \\
         \rowcolor{moongrey}
         Moon $\left(\SI{1.62}{\meter \per \second^2}\right)$ & \blueminus & \blueminus & \redcross & \greencheck \\
        \rowcolor{superblue}
         Super-Earth $\left(\SI{19.62}{\meter \per \second^2}\right)$ & \blueminus & \blueminus & \blueminus & \blueminus \\
        \midrule
         \multicolumn{5}{c}{\textbf{Base Pose Task - Power Consumption (W)}} \\
         \midrule
         \rowcolor{earthgreen}
         Earth $\left(\SI{9.81}{\meter \per \second^2}\right)$ & 59.8 & 59.8 & \textbf{24.3} & \textbf{24.3} \\
         \rowcolor{marsred}
         Mars $\left(\SI{3.73}{\meter \per \second^2}\right)$ & 16.0 & 19.6 & 7.3 & \textbf{5.0} \\
         \rowcolor{moongrey}
         Moon $\left(\SI{1.62}{\meter \per \second^2}\right)$ & 3.4 & 4.4 & 5.9 & \textbf{2.2} \\
        \rowcolor{superblue}
         Super-Earth $\left(\SI{19.62}{\meter \per \second^2}\right)$ & 89.9 & 111.8 & \textbf{31.4}  & 83.0 \\
        \midrule
    \end{tabular}
    \caption{Simulation evaluation of gravity-scaled and non-scaled locomotion and base pose policies across different gravity levels. We
commanded a walking speed of 0.4 m/s on flat terrain to assess the locomotion controllers. We qualitatively assessed the policies as good (\greencheck), medium (\blueminus), and bad (\redcross) and calculated the power consumption required as calculated by our power model (Sec.~\ref{fig:powertrain_losses}). The standard deviation of the velocity tracking error was less than \SI{0.05}{\meter \per \second} for all locomotion tasks.}
    \label{tab:gravity_scaling_results}
\end{table*}

To validate the gravity-based reward scaling (Sec.~\ref{sec:gravity_scaling}), we evaluated the baseline policies and the power-optimized policies for locomotion and pose control with and without gravity scaling across multiple gravity levels in simulation (Tab.~\ref{tab:gravity_scaling_results}). We conducted a qualitative visual assessment of the policies and measured their power consumption. We commanded a walking speed of \SI{0.4}{\meter \per \second} on flat terrain to assess the controllers. For the base pose controllers, the robot started at a height of \SI{0.32}{m}, and we then sequentially commanded a flat pitch, a maximum pitch of \SI{0.5}{rad}, and a maximum yaw of \SI{0.5}{\radian \per \second}. Each experiment lasted \SI{15}{\second} so we could average over multiple gait cycles and pose tracking motions. Tab.~\ref{tab:gravity_scaling_results} shows the results for all gravity levels, regularizations, and tasks. 

\subsection{Real-World Power Consumption Comparison in Multiple Gravity Environments}

\begin{table}[ht]
    \centering
    \begin{tabular}{c|c|c}
         \toprule
         \multicolumn{3}{c}{\textbf{Locomotion Task - Power Consumption (W)}} \\
         \midrule
         \midrule
         & \textbf{Baseline} & \textbf{Power-Optimized} \\
         \midrule
         \rowcolor{earthgreen}
         Earth $\left(\SI{9.81}{\meter \per \second^2}\right)$ & 30.4 & \textbf{23.4}  \\
         \rowcolor{moongrey}
         Moon $\left(\SI{1.62}{\meter \per \second^2}\right)$ & 19.2 & \textbf{12.2} \\
        \bottomrule
    \end{tabular}
    \caption{Real robot power consumption of the baseline and power-optimized locomotion controllers in Earth and lunar gravity. Note that we subtracted the robot's base consumption of \SI{77}{\watt}.}
    \label{tab:power_loco_real}
\end{table}

We deployed the Earth and Moon locomotion policies on the real robot using the gravity offload system presented in Sec.~\ref{sec:gravity_offload} to simulate lunar gravity. Note that this environment differs from actual lunar gravity, as the offload only provides a single force on the robot's base instead of offloading all links. The legs are instead internally offloaded as described in Sec.~\ref{sec:mdp}. The robot operated on a flat surface with high-friction mats. 
We compare the power consumption of the gravity-scaled baseline locomotion policy and the gravity-scaled power-optimized locomotion policy in Earth gravity and the lunar test environment (Tab.~\ref{tab:power_loco_real}). We powered the robot via an external power supply and measured the power supply voltage and current with an oscilloscope to measure the total power consumption. The robot's standby consumption - with all systems running but no torque applied to the drives - is \SI{77}{\watt}. We evaluated only the added consumption when running the control policies, which arises from the losses in the drivetrain (winding and gearbox losses) and the mechanical power at the joints (Sec.\ref{sec:power_reward}). For each run, we averaged the power over \SI{15}{\second}, which corresponds to roughly 22 gait cycles.

\section{DISCUSSION}
Across both tasks and all gravity levels, the power-optimized reward led to more power-efficient and qualitatively better policies than the baseline reward. 
Without gravity scaling, the power-optimized reward set does not scale well to other gravity levels, as it leads to erratic behavior, for example, very large and fast leg motions, when retraining for low gravity. The gravity scaling produces functional policies with the power-optimized reward set across all gravity levels. Furthermore, these rescaled policies are generally very power-efficient. The improved power efficiency is mainly due to the high base position, which means the legs are close to the singularity and require less torque to support the base~\cite{kolvenbach2021martianslopes}. In the Super-Earth case, the non-scaled power-optimized reward outperformed the gravity-scaled power-optimized reward. Based on this result, the scaling law should be revisited and tested across a wider range of gravity environments.
Scaling affected the baseline policies less than the power-optimized policies. However, the baseline policies were also generally less efficient. For each gravity and task, the best power-optimized policy outperformed the best baseline policy in qualitative and quantitative evaluations. 

In the real-world experiments, the power-optimized locomotion policy consumed less power than the baseline policy, which aligns with the simulation results. The relative difference between baseline and power-optimized was 23 \% in Earth gravity and 36 \% in the lunar test setup. 

We note that the standby power consumption of the robot due to the onboard computer, router, and motor controllers is six times larger than the power consumption of the efficient lunar locomotion policy. This relation shows that for a lunar mission with a legged robot, we need to optimize not only the control policy for power efficiency but also all onboard systems, especially the motor controllers.

While we trained our controllers on rough terrain with domain randomization on the ground parameters, we limited the evaluation of the power consumption to flat terrain for a fair comparison. We expect a similar qualitative trend between the power-optimized and the baseline policies on rough, granular terrain, but this hypothesis will need to be validated through further experiments.

The passive spring-loaded test setup allowed us to validate the lunar gravity policies on the real robot. While the policies transferred successfully and the behavior is visually comparable between the simulation and the real robot, additional work is needed to reduce the horizontal disturbance force. Furthermore, a tuning mechanism to match the offload force to the robot would be beneficial.

\section{CONCLUSION AND FUTURE WORK}
We have introduced and successfully validated a method for developing energy-efficient controllers for a legged robot across multiple tasks and gravity levels.

We have shown that using power-optimized reward functions leads to power-efficient policies for two tasks, namely locomotion and base pose control. However, naively transferring the same power-optimized rewards to different gravity leads to poor performance. Scaling the power rewards with gravity alleviates this issue and leads to functional and power-efficient controllers across multiple gravity environments. However, we saw that in the high-gravity Super-Earth case, the gravity scaling deteriorates the power efficiency. Future work should investigate scaling laws that remain efficient in high gravity levels. 

Additionally, we focused on the gravity scaling of locomotion policies here. In the future, tests with a gravity offload system 
on granular terrain are required to understand how the low-gravity locomotion is affected on such terrain.

Furthermore, we will modify the spring-loaded test setup to allow for fine-tuning of the offload force, which will allow accurate gravity offloading for multiple robots of similar mass.

\section*{ACKNOWLEDGMENTS}
We thank the \textit{Magnecko} team for their help with the experiments, the maintenance of the robot's software and hardware, and generally for building a well-engineered and easy-to-use system. We thank Filip Bjelonic for fruitful discussions on power modeling in robotic drivetrains.

\bibliographystyle{IEEEtran}
\bibliography{references}
\end{document}